\documentclass[letterpaper, 10 pt, conference]{ieeeconf}  % Comment this line out if you need a4paper

\IEEEoverridecommandlockouts                              % This command is only needed if 
\usepackage[nolist,nohyperlinks]{acronym}

\usepackage{graphicx}
\usepackage{subfigure}
\usepackage{subcaption}
\usepackage{amsmath}

\usepackage{tabularx}
\usepackage{booktabs}
\usepackage{multirow}
\usepackage{tikz}
\usepackage{amssymb}
\usepackage{pifont}
\usepackage{xcolor}

\usepackage{datetime}
\usepackage{subcaption}
\usepackage{microtype}

\title{\LARGE \bf
MarineGym: Accelerated Training for Underwater Vehicles with High-Fidelity RL Simulation
}

\author{Shuguang Chu$^{1}$, Zebin Huang$^{2,3}$, Mingwei Lin$^{1}$, Dejun Li$^{1}$ and Ignacio Carlucho$^{3}$% <-this % stops a space
\thanks{$^{1}$The State Key Laboratory of Fluid Power and Mechatronic Systems, Zhejiang University, Hangzhou, China}%
\thanks{$^{2}$School of Informatics, University of Edinburgh, Edinburgh, UK}%
\thanks{$^{3}$School of Engineering and Physical Sciences, Heriot-Watt University, Edinburgh, UK}%
\thanks{Mingwei Lin and Dejun Li are corresponding authors.(e-mail:lmw@zju.edu.cn;li\_dejun@zju.edu.cn;}
}

\begin{document}
\begin{acronym}
    \acro{RL}{reinforcement learning}
    \acro{USV}{Unmanned Surface Vehicle}
    \acro{UAV}{Unmanned Aerial Vehicle}
    \acro{AUVs}{Autonomous Underwater Vehicles}
    \acro{DRL}{Deep Reinforcement Learning}
    \acro{UV}{Underwater Vehicle}
    \acro{ROV}{Remotely Operated Vehicle}
    \acro{UUVs}{Unmanned Underwater Vehicles}
\end{acronym}

\maketitle
\thispagestyle{empty}
\pagestyle{empty}

\section{INTRODUCTION}
\ac{UUVs} have evolved to play a crucial role in ocean exploration. However, the nonlinear dynamics of \ac{UUVs} make achieving autonomy challenging in variable underwater environments. Learning-based methods, especially \ac{RL}, offer a promising solution by enabling robots to learn behaviours through trial and error to maximize predefined rewards \cite{kober_2013_ReinforcementLearningRobotics}. Recent \ac{RL} advancements have significantly enhanced control tasks for \ac{UUVs}, such as station-keeping \cite{walters_2018_OnlineApproximateOptimal}, target tracking \cite{masmitja_2023_DynamicRoboticTracking}, and trajectory tracking \cite{cui_2017_AdaptiveNeuralNetwork}.

However, applying \ac{RL} algorithms in underwater robotics is challenging due to complex environments and high risks. Therefore, simulation environments are crucial for developing learning-based methods for \ac{UUVs}. They provide a safe setting for extensive testing and training and allow for rigorous validation before transferring to real-world applications.

Current underwater robotic simulators exhibit significant limitations when interfacing with \ac{RL} methodologies. On the one hand, RL compatibility has only been introduced in a few recent works, such as HoloOcean \cite{potokar_2022_HoloOceanUnderwaterRobotics} and UNav-Sim \cite{amer_2023_UNavSimVisuallyRealistic}. On the other hand, there is a notable absence of research focusing on the training efficiency of \ac{RL} in these environments. Therefore, there is an emerging need for a specialized simulation platform that prioritizes robot learning and training efficiency in underwater robotics.

This extended abstract introduces a new UUV simulation framework, MarineGym, developed based on Isaac Sim \cite{liang2018GPUAcceleratedRoboticSimulationa}. It was designed to enhance \ac{RL} training efficiency using GPUs. Compared to existing simulators, MarineGym offers a 10,000-fold performance acceleration on a single GPU relative to real-time, thus reducing training time to just a few minutes. The extended abstract also provides four different tasks to demonstrate its effectiveness for training the \ac{RL} algorithm.

To summarise, this framework contributes:
\begin{itemize}
\item Accurately replicating the physical environment, encompassing physical laws, kinematics, and dynamics.
\item Supporting the parallel execution of multiple environment instances. This capability not only boosts training efficiency but also increases sample diversity, helping to prevent overfitting.
\item Ensuring compatibility with existing \ac{RL} frameworks (such as TorchRL) and offering user-friendly APIs to facilitate seamless integration and usage.
\end{itemize}

\begin{figure*}
\centering
\includegraphics[width=1.0\textwidth]{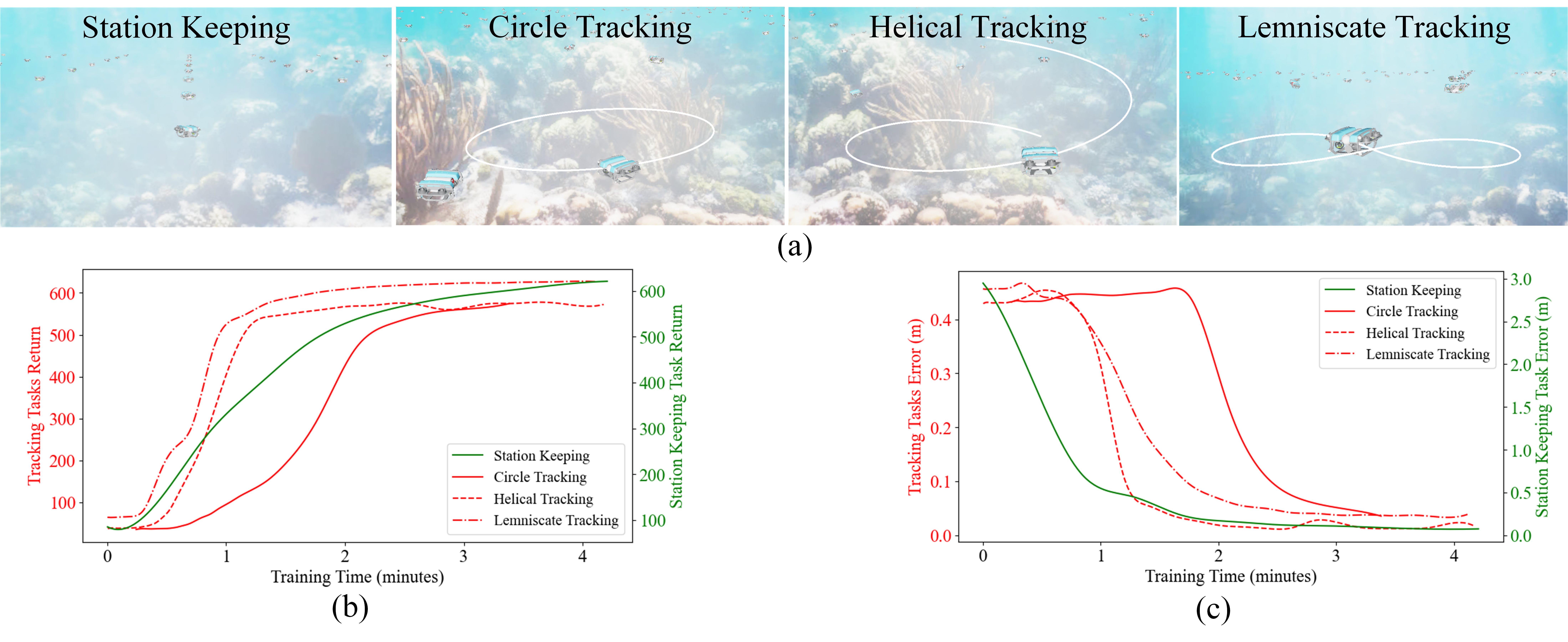}
\caption{The results of training four different tasks in MarineGym. (a) Rendered simulation scene during the validation process of the trained agent. (b) Return over training time. (c) Position error over training time.}
\label{Fig:training results}
\end{figure*}

\section{METHOD}
MarineGym comprises four main components: (1) the UUV Dynamics Simulation Module, featuring classic UUV models like BlueROV2 and BlueROV2 Heavy for dynamic simulations and physical parameter domain randomization; (2) the Physical Scene Simulation Module, utilizing NVIDIA RTX technology for real-time underwater scene rendering and realistic sensor data generation; (3) the \ac{RL} Environment Module, offering a variety of tasks for customized simulation and training; and (4) the RL and Deep Learning Interface, incorporating toolkits such as PyTorch and TorchRL, enabling efficient development and optimization of learning algorithms.

The hydrodynamic calculations of the UUV are based on Fossen's equation of motion \cite{fossen2011handbook}, a well-known theoretical method for UUV simulation. To adapt it into the framework, the equation is extended as follows:
\vspace{-0.5em}
\begin{equation}
\begin{aligned}
\mathbf{\tau} = & \underbrace{\mathbf{M}_{RB}\dot{\mathbf{\nu}} + \mathbf{C}_{RB}(\mathbf{\nu})\mathbf{\nu} + \mathbf{g}_{RB}(\mathbf{\eta})}_{\text{Rigid-body term}} \\
& + \underbrace{\mathbf{M}_A\dot{\mathbf{\nu}} + \mathbf{C}_A(\mathbf{\nu})\mathbf{\nu} + \mathbf{D}(\mathbf{\nu})\mathbf{\nu} + \mathbf{g}_A(\mathbf{\eta})}_{\text{Hydrodynamic term}}
\end{aligned}
\label{Equ:fossen}
\end{equation}
\vspace{-1em}

\noindent where $\mathbf{\eta}=\{x,y,z,\phi,\theta,\psi\}$ represents the vehicle's pose in the North-East-Down (NED) coordinate frame, and $\mathbf{\nu}=\{u,v,w,p,q,r\}$ represents the vehicle's velocity in the body-fixed frame. The rigid-body term includes inertia matrix $\mathbf{M}_{RB}$, Coriolis and centripetal matrix $\mathbf{C}_{RB}(\mathbf{\nu})$, and gravity $\mathbf{g}_{RB}(\mathbf{\eta})$. The hydrodynamic term includes added-mass matrix $\mathbf{M}_A$, added-mass Coriolis and centripetal matrix $\mathbf{C}_A(\mathbf{\nu})$, damping matrix $\mathbf{D}(\mathbf{\nu})$, and buoyancy $\mathbf{g}_A(\mathbf{\eta})$.

There are multiple sub-simulation steps within a single simulation step. Each sub-simulation step updates the dynamics information according to the Equation \ref{Equ:fossen}. The rigid-body term is automatically handled by PhysX physics engine, while the hydrodynamic term is computed by the rapid parallel hydrodynamics calculation module implemented in this study. This term is then applied as external forces to the UUV's body, thereby updating its motion states. 

\section{EXPERIMENTS}

To validate the simulator's performance, four distinct control tasks using the UUV BlueROV2 Heavy were designed: (1) station-keeping, (2) circle tracking, (3) helical tracking, and (4) lemniscate tracking. The station-keeping task aims to position the ROV at a specific point and maintain stability. The tracking tasks aim to control the ROV to follow a predefined trajectory. The circle trajectory is a circular path in a two-dimensional plane, while the helical trajectory adds a vertical component to the circular path. The lemniscate trajectory features complex curvature changes.

The state space includes the motion information of UUV and the relative position to the target, defined as:

\vspace{-1em}
\begin{equation}
\begin{cases}
\{\mathbf{\eta}^e, \mathbf{\nu}\} & \text{Task}_{1} \\
\{\mathbf{\eta}_1^e, \mathbf{\eta}_2^e, \ldots, \mathbf{\eta}_m^e, \mathbf{\nu}\} & \text{Task}_{2,3,4}
\end{cases}
\end{equation}
\vspace{-1em}

In the station-keeping task, the state space includes the current position error relative to target $\mathbf{\eta}^e=\{x^e,y^e,z^e,\phi^e,\theta^e,\psi^e\}$ and the current velocities $\mathbf{\nu}$. For the tracking tasks, the state space includes the error between the current position and subsequent $m$ trajectory points, indicated by $\{\mathbf{\eta}_1^e, \mathbf{\eta}_2^e, \ldots, \mathbf{\eta}_m^e\}$. 

The action space is defined as:
\vspace{-0.5em}
\begin{equation}
\{ f_i \mid i \in [1, N] \}
\end{equation}
where $f_i$ represents the throttle setting for the $i$-th thruster, and $N$ is the total number of thrusters.

\vspace{-1em}
\begin{equation}
\scriptsize
\begin{cases}
\sqrt{\left(x-x_r\right)^2+\left(y-y_r\right)^2+\left(z-z_r\right)^2} & \text{Task}_{1} \\
\sum_{i=1}^n \sqrt{\left(x_i-x_{r, i}\right)^2+\left(y_i-y_{r, i}\right)^2+\left(z_i-z_{r, i}\right)^2} & \text{Task}_{2,3,4}
\end{cases}
\end{equation}
\vspace{-1em}

\noindent where $\{x_r,y_r,z_r\}$ represents the target position in station-keeping task, and $\{x_{r,i},y_{r,i},z_{r,i}\}$ represents the $i$-th target point position on the trajectory in tracking task.

All tasks are designed in a Gym-style format to enable integration with standard RL libraries. We used the PPO algorithm, implemented in PyTorch with GPU acceleration, for training the tasks. Both the simulation and the algorithm were executed on an NVIDIA GeForce RTX 3060, which reduced CPU-GPU data transfers. The experimental results are shown in Fig.\ref{Fig:training results}. Leveraging the proposed framework, we achieved a sampling rate of over 700,000 steps per second, corresponding to a 10,000-fold acceleration relative to real-time. All four tasks completed training within four minutes. The positional error for the station-keeping task was less than 0.1 meters, and the tracking error for the three trajectory tracking tasks did not exceed 0.02 meters.

This abstract presents a UUV framework specifically designed for \ac{RL}. The framework utilizes GPU for large-scale sampling and supports domain randomization. Future work includes integrating more UUV models and sensor support, and validating the performance of trained agents after Sim2Real transfer. This framework is expected to significantly advance the development of \ac{RL} in UUV and to enhance their overall intelligence capabilities.

\bibliographystyle{ieeetr}

\bibliography{IEEEabrv, reference.bib}

\begin{thebibliography}{1}

\bibitem{kober_2013_ReinforcementLearningRobotics}
J.~Kober, J.~A. Bagnell, and J.~Peters, ``Reinforcement learning in robotics: A survey,'' {\em The International Journal of Robotics Research}, vol.~32, pp.~1238--1274, Sept. 2013.

\bibitem{walters_2018_OnlineApproximateOptimal}
P.~Walters, R.~Kamalapurkar, F.~Voight, E.~M. Schwartz, and W.~E. Dixon, ``Online approximate optimal station keeping of a marine craft in the presence of an irrotational current,'' {\em IEEE Transactions on Robotics}, vol.~34, pp.~486--496, Apr. 2018.

\bibitem{masmitja_2023_DynamicRoboticTracking}
I.~Masmitja, M.~Martin, T.~O'Reilly, B.~Kieft, N.~Palomeras, J.~Navarro, and K.~Katija, ``Dynamic robotic tracking of underwater targets using reinforcement learning,'' {\em Science Robotics}, vol.~8, p.~eade7811, July 2023.

\bibitem{cui_2017_AdaptiveNeuralNetwork}
R.~Cui, C.~Yang, Y.~Li, and S.~Sharma, ``Adaptive neural network control of auvs with control input nonlinearities using reinforcement learning,'' {\em IEEE Transactions on Systems, Man, and Cybernetics: Systems}, vol.~47, pp.~1019--1029, June 2017.

\bibitem{potokar_2022_HoloOceanUnderwaterRobotics}
E.~Potokar, S.~Ashford, M.~Kaess, and J.~G. Mangelson, ``Holoocean: An underwater robotics simulator,'' in {\em 2022 International Conference on Robotics and Automation (ICRA)}, pp.~3040--3046, IEEE, May 2022.

\bibitem{amer_2023_UNavSimVisuallyRealistic}
A.~Amer, O.~{\'A}lvarez-Tu{\~n}{\'o}n, H.~{\.I}. U{\u g}urlu, J.~Le~Fevre~Sejersen, Y.~Brodskiy, and E.~Kayacan, ``Unav-sim: A visually realistic underwater robotics simulator and synthetic data-generation framework,'' in {\em 2023 21st International Conference on Advanced Robotics (ICAR)}, pp.~570--576, IEEE, Dec. 2023.

\bibitem{liang2018GPUAcceleratedRoboticSimulationa}
J.~Liang, V.~Makoviychuk, A.~Handa, N.~Chentanez, M.~Macklin, and D.~Fox, ``Gpu-accelerated robotic simulation for distributed reinforcement learning,'' in {\em Proceedings of The 2nd Conference on Robot Learning}, pp.~270--282, PMLR, Oct. 2018.

\bibitem{fossen2011handbook}
T.~I. Fossen, {\em Handbook of marine craft hydrodynamics and motion control}.
\newblock John Wiley \& Sons, 2011.

\end{thebibliography}

\end{document}